\RequirePackage{fix-cm}

\documentclass[twocolumn]{svjour3}          
\smartqed  

\usepackage{graphicx}
\usepackage{graphicx}
\usepackage{xcolor}
\usepackage{amssymb}
\usepackage{commath}
\usepackage{pgfplots}
\usepackage{subcaption}
\newcommand{\cmmnt}[1]{}
\usepackage{smartdiagram}
\usepackage{amsmath,amsfonts,amssymb}
\usepackage[ruled,vlined,linesnumbered]{algorithm2e}
\usepackage{lipsum}
\usepackage{mathtools}
\usepackage{cuted}
\usepackage{balance}
\usepackage{textcomp}
\usepackage{latexsym}
\begin{document}

\title{Classification of Chinese Handwritten Numbers with Labeled Projective Dictionary Pair Learning
}


\author{Rasoul Ameri \and Ali Alameer \and Saideh Ferdowsi \and Kianoush Nazarpour \and Vahid Abolghasemi 
}


\institute{Rasoul Ameri \at Shahrood University of Technology, Iran 
           \and
           Ali Alameer \at Queen's University Belfast, UK 
            \and
              Saideh Ferdowsi \at University of Essex, UK
               \and            
              Kianoush Nazarpour \at University of Edinburgh, UK
               \and
              Vahid Abolghasemi \at University of Essex, UK\\
              \email{v.abolghasemi@essex.ac.uk}
}

\date{Received: date / Accepted: date}

\maketitle

\begin{abstract}
Dictionary learning is a cornerstone of image classification. We set out to address a longstanding challenge in using dictionary learning for classification; that is to simultaneously maximise the discriminability and sparse-representability power of the learned dictionaries. Upon this premise, we designed class-specific dictionaries incorporating three factors: discriminability, sparsity and classification error. We integrated these metrics into a unified cost function and adopted a new feature space, i.e., histogram of oriented gradients (HOG), to generate the dictionary atoms. The rationale of using HOG features for designing the dictionaries is their strength in describing fine details of crowded images. The results of applying the proposed method in the classification of Chinese handwritten numbers demonstrated enhanced classification performance $(\sim98\%)$ compared to state-of-the-art deep learning techniques (i.e., SqueezeNet, GoogLeNet and MobileNetV2), but with a fraction of parameters. Furthermore, combination of the HOG features with dictionary learning enhances the accuracy by $11\%$ compared to the case where only pixel domain data are used. These results were supported when the proposed method was applied to both Arabic and English handwritten number databases.
\keywords{Deep Learning \and Dictionary Learning \and Handwritten number recognition \and Image Classification \and Sparse Coding}
\end{abstract}

\section{Introduction}
Handwritten number recognition has remained a challenging research topic within pattern recognition \cite{10.3389/fncom.2015.00099}, still attracting many researchers \cite{qiao2018adaptive,9110900,farsiNumber}. A generic handwritten recognition system uses machine learning to interpret and recognise the received handwritten data from different sources, e.g. emails, bank cheque, papers, images, etc. Traditional recognition systems comprise two major stages: feature extraction and classification. The first stage transforms the input data into a space to accurately describe the data while reducing the amount or dimensionality of data, and the second stage assigns the input data to an associated class. Various techniques have been proposed for handwritten numbers classification where the main challenge is to learn efficient and comprehensive model capable of handling a wide diverse range of handwritten styles. A brief overview of the handwritten number recognition literature with focus on two main learning-based methods, i.e., deep learning and dictionary learning, as well as HOG-related techniques is provided in the following.

Researchers proposed a classifier based on LeNet-5 and support vector machine (SVM) for handwritten number recognition \cite{yu2015handwritten,lauer2007trainable}. Other relevant studies \cite{choudhury2018handwritten}, involved utilising histogram of oriented gradients (HOG) and SVM for feature extraction and classification, respectively. In a recent relevant work \cite{9092067}, HOG and Gabor filter were used as descriptors for feature extraction from Arabic words, leading to promising results using a k-nearest neighbor (kNN) classifier. HOG focuses on the structure of an object and can extract gradient and orientation of edges in a given image. It was initially proposed for human detection, however, it has recently shown great influence for feature extraction from handwritten numbers \cite{choudhury2018handwritten}. Nevertheless, the efficacy of this powerful feature descriptor has not been thoroughly studied in this context, particularly for Chinese handwritten numbers.

Recently, deep learning-based approaches have also been proposed to classify handwritten numbers. A deep unsupervised network was proposed in \cite{9110900} to learn invariant image representation from unlabeled data. The network architecture comprised a cascade of convolutional layers trained sequentially to represent multiple levels of features. A deep neural network classifier has been proposed in \cite{farsiNumber} for Farsi handwritten phone numbers recognition. In another work \cite{9084035}, Bengali handwritten number detection was performed using a deep structure called region proposal networks (RPN). Despite the promising performance of deep learning-based approaches, they require a large training dataset with numerous parameters to be tuned. Conventional methods, in comparison, are more appropriate for images with lower resolution, such as, handwritten characters \cite{luo2019multi}. 
Moreover, in spite of several works on Chinese handwritten \textit{characters} recognition \cite{8563234,9085435,8869841}, there are no reported works on performance of deep learning for Chinese handwritten \textit{numbers}. 

Dictionary learning is another learning-based approach that has achieved promising results in image classification \cite{DLbook/3265797}. The key success of this approach originates from the fact that a sample from a class of interest can be efficiently represented as a sparse linear combination of other samples of the same class \cite{wright2008robust}. Most conventional dictionary learning methods involve two major steps in which their performances are highly interdependent; sparse coding and dictionary update. The quality of obtained dictionary is crucial for generating a sufficient sparse representation, e.g., sparsity and grouping. Therefore, some dictionary learning methods attempt to adaptively design dictionaries and to efficiently represent the input data, e.g., gray-scale images \cite{huang2007sparse}. A naive way of building dictionaries is to stack all the training data into a matrix (the so-called dictionary), however, this approach leads to huge and redundant dictionaries; impractical to be stored or utilised for any purposes. Thus, many studies utilise machine learning techniques for obtaining dictionaries by extracting low-dimensional features domain from the given training data. The main aim of such techniques is to obtain a dictionary with approximately independent atoms that convey formative information of the input data. The utilised learning process, however, depends on the structure and nature of the input images too. The efficiency of a dictionary is also dependent on the total number of coefficients contained in the associated sparse vectors \cite{aharon2006k}. These sparse vectors together with dictionary atoms act as a coder for best approximating the data of interest. This idea can be extended from data representation to data classification, i.e., learning class-specific dictionaries. 

Dictionary learning has been applied for face recognition \cite{wright2008robust} and brain signal classification \cite{shin2012sparse}. A well-established work on dictionary learning for image classification is the sparse representation classifier (SRC) \cite{wright2008robust}. This technique uses sparse representation and learned dictionaries for classification of images in pixel domain. It is increasingly being extended and used for a wide variety of image analysis, representation and classification tasks. A recently developed supervised dictionary learning approach constructs image classes using a shared dictionary and discirmintave class models \cite{10.5555/2981780.2981909}.
A limitation associated with the above approach is that the size of the dictionary increases when adding more classes and that degrades the classification performance. To scale to large training sets, researchers have proposed methods that learn a dictionary by merging its atoms by optimising a predefined objective function  \cite{fulkerson2008localizing,winn2005object}. This mechanism decreases the mutual information between the dictionary atoms and the class labels \cite{fulkerson2008localizing}. Additionally, it minimises the loss of mutual information between the histogram of dictionary atoms over signal constituents \cite{winn2005object}. Despite the acceptable performance of these dictionaries, they tend to be computationally expensive due to the feature merging stage. 
Other approaches involve jointly learning the dictionary and classifier using an optimised objective function. K-singular value decomposition (K-SVD) method was utilised to train the dictionary \cite{aharon2006k}. This method has been applied to a variety of image processing problems, including inpainting missing pixels and image compression. The authors in \cite{zhang2010discriminative}, proposed a method for dictionary learning that jointly learns the classifier parameters and dictionary for face recognition. A method called label consistent K-SVD (LC-KSVD) was proposed to learn a discriminative dictionary for sparse coding \cite{jiang2011learning}, \cite{jiang2013label}. Introducing labels and classification error to the objective function has leveraged the performance of LC-KSVD method.
	\begin{figure*}[ht!]
		\centering
		\includegraphics[scale=.55, trim=1cm 6cm 1cm 6cm]{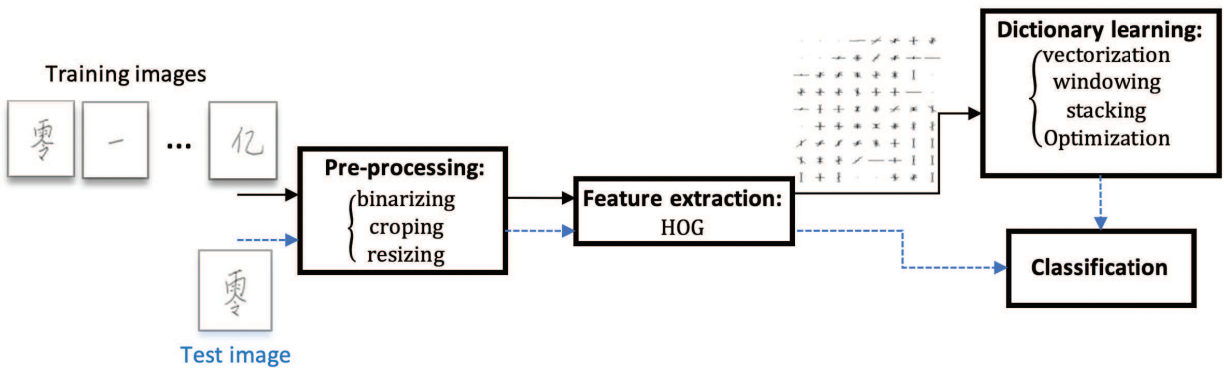}
		\caption{Block diagram of the proposed method. Black solid and blue dashed lines, respectively, illustrate the flow of training and testing phases. }
		\label{Fig_blk}
	\end{figure*}

Traditional dictionary learning methods merely rely on \textit{synthesis} dictionaries in which the input data is in a sparse latent subspace. Synthesis dictionaries can well preserve the local structures of the data. In contrast, \textit{analysis} dictionaries, which was recently introduced, rely on the assumption that the input data can be transformed into a latent sparse subspace by its corresponding dictionary \cite{8745700}. Analysis dictionary can produce sparse representation of data via simple data transformation, i.e. linear projection (simple dot product) without applying $\ell_0$/$\ell_1$ minimisation, which are considered computationally expensive operations due to their non-convex nature. For instance, an analysis discriminative dictionary learning has been proposed in \cite{8038251} to process two-dimensional images. The method imposes a sparse $\ell_{2,1}$-norm constraint on the coding coefficients and attempts to learn dictionaries, representations, and linear classifiers as discriminant as possible. Recently, projective dictionaries pair learning (DPL) was proposed where both types of dictionaries, i.e. analysis (for generating discriminative code by linear projection) and synthesis (for reconstructing the data) were used \cite{gu2014projective}. It benefits from an analysis-synthesis dictionary pair that avoids the need for utilising $\ell_{0}$-norm or $\ell_{1}$-norm minimisation. DPL has shown promising performance on face recognition application over state-of-the-art techniques. We recently applied this method for classification of brain activities in a brain-computer interfacing application \cite{ameri2016projective}. In addition, we proposed an extension to the DPL method, called incoherent dictionary pair learning (InDPL), for the classification of Chinese handwritten numbers \cite{abolghasemi2018incoherent}. InDPL adds a dictionary incoherence penalty to the DPL cost function in order to increase the discriminability and thus improving the classification performance. 

Some recent works have addressed combination of dictionary learning and deep learning. Deep dictionary learning was proposed in \cite{7779008} for building deeper architectures using the layers of dictionary learning. A method called deep micro-dictionary learning and coding network was proposed in \cite{8658671} which includes most of the standard deep learning layers (pooling, fully, connected, input/output, etc.). However, the deep learning architecture is augmented by replacing fundamental convolutional layers with a novel compound dictionary learning and coding layers.  In \cite{1446833}, scalability and speed of deep learning were combined with dictionary learning to significantly reduce the number of parameters. This convolutional dictionary learning based auto-encoder was proposed for natural exponential-family distributions such as image denoising and neural spiking data analysis.

In this paper, we further extend the DPL method and present a Chinese handwritten numbers recognition system that exploits the class labels information within the minimisation process. In our proposed labeled projective DPL (LpDPL), both synthesis and analysis dictionaries as well as class labels are used to calculate the sparse codes. A solution based on alternative minimisation is proposed that provides optimum trade-off between sparsity and grouping effect without using $\ell_0$ and $\ell_1$ regularisers. Instead of using raw pixel domain information as input to dictionary learning stage, here, we propose to use HOG (histogram of oriented gradient) descriptors. To the best of our knowledge, this is the first work reporting the use of HOG features in the context of dictionary learning. The motivation behind opting HOG is its robustness demonstrated in digit/character recognition applications as reviewed earlier in this section. Here, we extract and embed the HOG-based features into dictionary learning process. In addition, we mathematically utilise the classification labels information and propose a novel cost function for simultaneous dictionary learning and discrimination. Finally, to quantitatively evaluate performance, we apply several deep neural network architectures in addition to other well-established dictionary learning methods, and further compare the efficiency of two learning-based methods, i.e.``dictionary learning'' and ``deep learning'' methods. 

\begin{figure*}[t!]
	\centering
	\begin{subfigure}[A]{\textwidth}
			\centering
			\includegraphics[scale=.45]{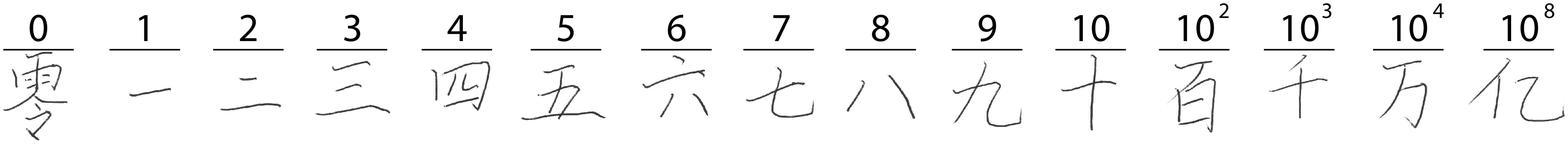}
			\caption{Simplified Chinese handwritten numbers.}
			\label{Fig_01A}
	\end{subfigure}%
	\newline
	\begin{subfigure}[B]{\textwidth}
			\centering
			\includegraphics[scale=.45]{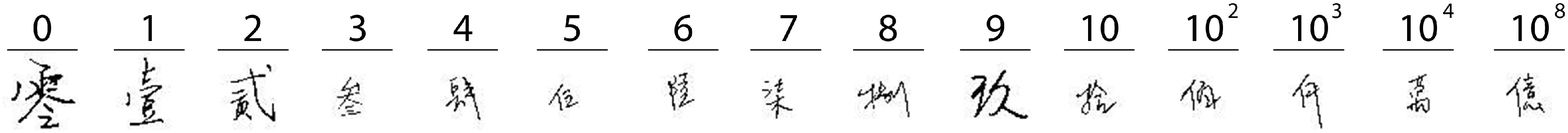}
			\caption{Traditional Chinese handwritten numbers.}
			\label{Fig_01B}
	\end{subfigure}
	\newline
	\begin{subfigure}[C]{\textwidth}
			\centering
			\includegraphics[scale=.45]{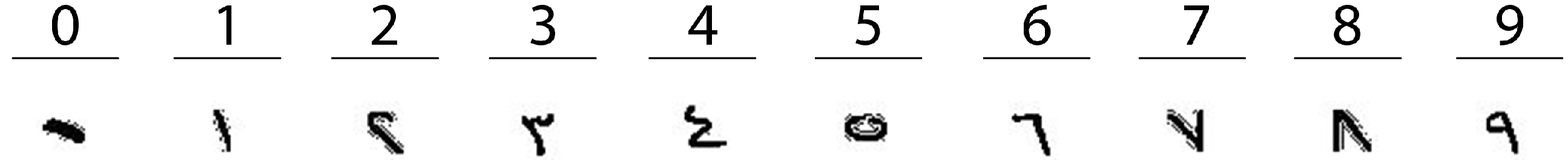}
			\caption{Arabic handwritten numbers.}
			\label{Fig_01C}
	\end{subfigure}	
	\newline
	\begin{subfigure}[C]{\textwidth}
		\centering
		\includegraphics[scale=.45]{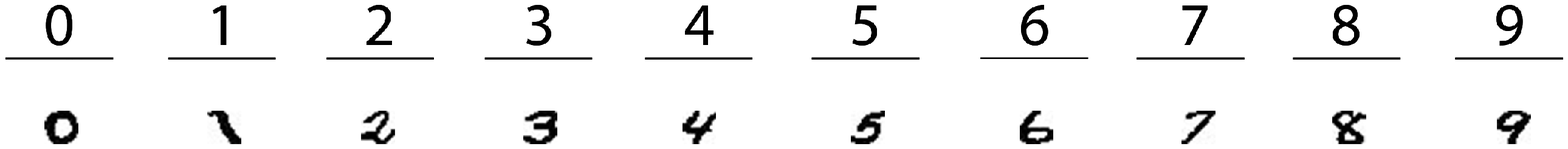}
		\caption{English handwritten numbers (USPS).}
		\label{Fig_01D}
	\end{subfigure}
	\caption{Sample images from three different handwritten numbers databases and their equivalent English numbers.}
\end{figure*}

\section{Materials and methods}
The proposed method comprises four stages: 1) preprocessing, 2) HOG feature extraction, 3) LpDPL execution, and 4) classification. In the first stage, we enhance the quality of collected images. Then, the HOG features are extracted by calculating the orientation histograms of edge intensity in local regions. In the third stage, the features of training samples construct the dictionary columns (atoms). Labeled projective dictionary pair learning extracts the features before the fourth stage, whereby Chinese numbers are classified. The block diagram of the proposed method representing different steps in training and testing phases is shown in Figure \ref{Fig_blk}.

\subsection{Image Databases}
We use two Chinese handwriting numbers databases to analyse the effectiveness of our method. The first is an open source database, which was published in our earlier work \cite{DataBase2019}. The database contains 15,000 handwritten numbers from 100 Chinese nationals studying at Newcastle University, UK. Each participant wrote the 15 numbers of Figure \ref{Fig_01A}, 10 times. Another independent Chinese handwritten numbers database, which consists of 5,100 handwritten numbers from 34 persons, was also used to analyse this method. Each person wrote 10 times the 15 numbers illustrated in Figure \ref{Fig_01B}.

In addition to the above Chinese databases, Arabic (MADBase \cite{AHD}) and English numbers databases (USPS \cite{Hull1994}) were considered as case studies. MADBase consists of 70,000 digits written by 700 persons that each person wrote 10 times each digit from 0-9. Similarly, USPS database consists of 7291 training samples and 2007 test samples of digits 0-9 in form of grayscale images. Sample images of these two databases are represented in Figures \ref{Fig_01C} and \ref{Fig_01D}.
		
\subsection{Image pre-processing}
\begin{figure}[t]
		\centering\includegraphics[scale=1]{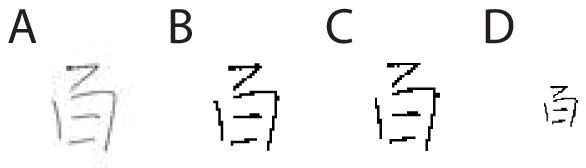}
		\caption{Example result of image pre-processing step for Chinese number 100. (A) the original grayscale image of size $64\times 64$; (B) binarised image using Otsu's method; (C) cropped image; (D) resized image. The images are made negative for ease of presentation.}
		\label{Fig_02}
	\end{figure}

The pre-processing phase includes two parts; image enhancement and noise removal. Initially, the scanned images (e.g. Figure \ref{Fig_02}A) are converted to grayscale, then, the global image threshold is determined by using Otsu's method \cite{otsu1979threshold} to convert the image from grayscale to binary, as shown in Figure \ref{Fig_02}B. The cropping operation is performed such that the number would be at the center of a predefined bounding box, for example Figure \ref{Fig_02}C. In the last step, images are re-sampled to $32\times 32$ pixels which is applied to equalise the dimensions of input images (Figure \ref{Fig_02}D).
	
\subsection{Feature extraction}
\begin{figure}[pt!]
		\centering
		\includegraphics[scale=.8]{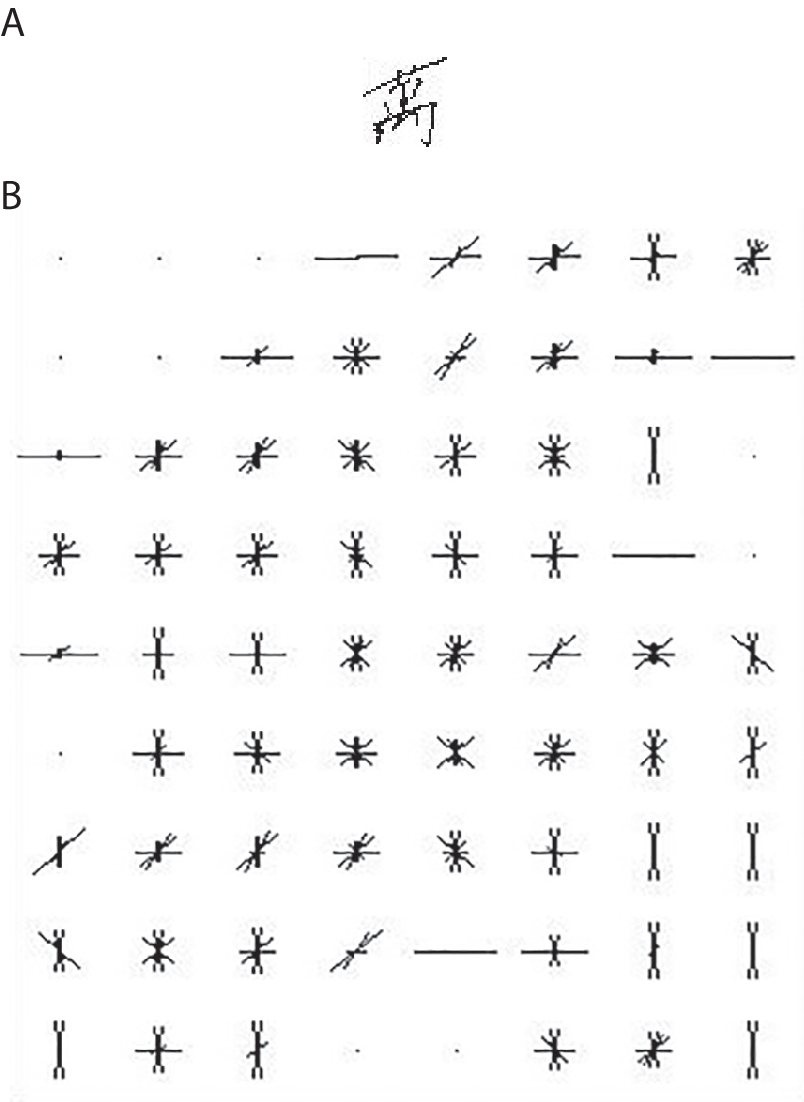}
		\caption{HOG feature extraction for Chinese handwriting number. (A) an example of input image number 0; (B) HOG features of the input image; identifying all possible directions and angles. }
		\label{Fig_03}
	\end{figure}

HOG counts occurrences of gradient orientation in pre-defined parts of an image \cite{Freeman1995OrientationHF}. We divided the input image into small square cells of size $3\times 3$. Then, the histogram of gradient directions based on the central differences is computed which is referred to as histogram of oriented gradient. We then normalise the local histograms based on the minimum and maximum image contrast. This is to enable the dictionaries to generalise to different variation conditions. Figure \ref{Fig_03} illustrates an example of this stage whereby HOG has identified all possible directions and angles (Figure \ref{Fig_03}B).

\subsection{Labeled projective dictionary pair learning}
DPL method learns a synthesis dictionary and an analysis dictionary jointly for classification \cite{gu2014projective}. In this paper, we enhance the DPL method in two ways. Firstly, we embed the HOG features into the system hierarchy to create more representative and discriminant dictionaries. The DPL method is customised for the purpose of Chinese handwritten numbers recognition. Secondly, we introduce the prior knowledge of the class labels as a new term in the proposed cost function. The proposed method is mathematically expressed as follows.

Let us define all the the input preprocessed images which are collated for training as $X=[X_{1},X_{2},\cdots, X_{Q}]$, where $X_{q}\in R^{n\times k}$ encompasses the samples of $q$-th class (of total $Q$ classes), $k$ is number of training samples for class $q$, and $n$ is the length of training vectors (vectorised image). We introduce analysis dictionary as $P \in R^{m \times (Q \times n)} $ where $m$ and $ Q \times n $ are number of rows and columns, respectively. The sparse coefficient matrix is expressed as $A = PX  $. Also, we developed synthesis dictionary $ D \in R^{n \times (Q \times m)} $ which will be explained further. Generally, a dictionary learning and classification expression is defined as:
\begin{equation}\label{equation.1}
	{<P^{*}, D^{*}> = \operatorname*{argmin}_{P,D} {\norm{X-DPX}_{F}^{2}}+\Psi(D,P,X)}
\end{equation}	
where the term $ {\norm{X-DPX}_{F}^{2}} $ denotes the reconstruction error. The crucial task here is to design an appropriate penalty function $\Psi$ that leads to a successful classification. We consider three important factors to successfully form the cost function with the following objectives: 1) obtaining a sparse representation of the coefficients $PX$; 2) learning class-specific dictionaries, and 3) minimising the classification error. In what follows, we propose a new design for $\Psi$ to meet the aforementioned criteria, i.e. having a discrimination power in addition to a minimising the classification error. Then, a recurrent alternating approach is proposed to minimising the proposed objective function and finding suitable dictionaries for each class.

Let us define $ P = \{P_{1},\cdots P_{q},\cdots,P_{Q}\}$ and $ D = \{D_{1}, \cdots, D_{q},\cdots,D_{Q}\} $ where $ P_{q}\in R^{m \times n} $ and $ D_{q} \in R^{n \times m} $ are respectively analysis and synthesis sub-dictionaries, corresponding to class $ q $. For discrimination power, the obtained analysis dictionary $ P_{q} $ should be only associated with class $ q $ and unrelated to other classes $ q' $. Mathematically, it should satisfy:
\begin{equation}\label{equation.2}
		{P_{q}X_{q'}\approx 0;q\ne q' \; \; \textrm{and} \; \; 1\leq q,q'\leq Q}.
\end{equation}
\noindent where $X_{q'}$ includes all samples but those from class $q$, and $P_{q}X_{q'}\approx 0$ means that the analysis dictionary associated to class $q$ should solely be able to represent samples from class $q$. Such discriminability can be reformulated by Frobenious norm as $\norm{P_{i}\overline{X_{i}}}_{F}^{2}$ and added to the reconstruction error in (\ref{equation.3}). The matrix $\overline{X_{i}}$ denotes the complementary data matrix to $ X_{i} $, meaning that it encompasses samples of all classes except those for $i$-th class:
\begin{eqnarray}\label{equation.3}
		\sum_{i=1}^{Q} {\norm{X_{i}-D_{i}P_{i}X_{i}}_{F}^{2}} 
		 + \lambda_{1}\norm{P_{i}\overline{X_{i}}}_{F}^{2}.
\end{eqnarray}
Although (\ref{equation.3}) can enforce the synthesis dictionaries to be discriminant, it does not utilise this feature for the analysis dictionaries. Since the class labels are available, we propose to add a linear predictive classifier $ f(X;W) = WX $ to (\ref{equation.3}) in order to enforce analysis dictionaries to provide a higher level of discrimination. This is to incorporate classification error term in the objective function. Let $ H $ be a label matrix for input samples $ X $, and $ W $ denotes classifier parameters. To estimate $P^{*}, D^{*}, W^{*}$ the following minimisation problem is proposed:
\begin{eqnarray}\label{equation.4}
		\operatorname*{argmin}_{P,D,W} &&\sum_{i=1}^{Q} {\norm{X_{i}-D_{i}P_{i}X_{i}}_{F}^{2}} 
		 + \lambda_{1}\norm{P_{i}\overline{X_{i}}}_{F}^{2} \nonumber\\ &&+ \lambda_{2}\norm{H_{i}-W_{i}P_{i}X_{i}}_{F}^{2}\nonumber\\ &&s.t.\  \norm{d_{j}}_{2}^{2}\le 1 \ \textrm{for} \ j=1,2,\cdots m
\end{eqnarray}
where $ H=\{H_{i},\cdots,H_{q},\cdots,H_{Q}\} $ and $ H_{q} \in R^{Q \times K} $ is the binary label matrix corresponding to an input sample $ X_{q} $. Also, $d_j$ refers to $j$-th column of the corresponding dictionary $D$. The below example shows the values of $H_2$ corresponding to four samples and three classes:
	\[H_{2} = \left[
	\begin{matrix}
		0 & 0 & 0 & 0\\
		1 & 1 & 1 & 1\\
		0 & 0 & 0 & 0
	\end{matrix}
	\right]. \]
	
Equation (\ref{equation.4}) is generally non-convex and cannot be simultaneously solved for all variables. However, if we replace $A=PX$ into (\ref{equation.4}), the objective function will be converted to (\ref{equation.5}), where $P^{*}, D^{*}, W^{*},$ and $A^{*}$ can be calculated using an alternate minimisation technique:
\begin{eqnarray}\label{equation.5}
		\operatorname*{argmin}_{P,D,W,A}&& \sum_{i=1}^{Q} {\norm{X_{i}-D_{i}A_{i}}_{F}^{2}} + \lambda_{1}\norm{P_{i}\overline{X_{i}}}_{F}^{2} \nonumber \\ 
		&&+ \lambda_{2}\norm{H_{i}-W_{i}A_{i}}_{F}^{2}+ 				\lambda_{3}\norm{P_{i}X_{i}-A_{i}}_{F}^{2}
		\nonumber\\
		&& s.t. ~ \norm{d_{j}}_{2}^{2}\le 1
\end{eqnarray}
where $ \lambda_{1} $, $ \lambda_{2} $ and $ \lambda_{3} $ are the regularisation parameters and are set empirically. For the optimisation, equation (\ref{equation.5}) can be alternated between the following steps.
		
\noindent \textbf{Step 1}: Fix $ D, W, P $ and update $ A $:	
				when fixing $ D, W, P $ to update $ A $, we omit the terms independence of $ A $ from \ref{equation.5}:
				\begin{eqnarray}\label{equation.6}
					A^{*} &=& \operatorname*{argmin}_{A} \sum_{i=1}^{Q} {\norm{X_{i}-D_{i}A_{i}}_{F}^{2}} \nonumber \\ &&+~\lambda_{2}\norm{H_{i}-W_{i}A_{i}}_{F}^{2}\nonumber \\
					&&+ ~			\lambda_{3}\norm{P_{i}X_{i}-A_{i}}_{F}^{2}.
				\end{eqnarray}
				This equation is convex and differentiable. After obtaining derivative with respect to $ A $ and equating it to zero, we have:
				\begin{eqnarray}\label{equation.7}
					A^{*} &=& (D_{i}^{T}D_{i}+W_{i}^{T}W_{i}+\lambda_{3}I)^{-1}\nonumber\\ &&(D_{i}^{T}X_{i}+\lambda_{2}W_{i}^{T}H_{i}+\lambda_{3}P_{i}X_{i})
				\end{eqnarray}
				
\noindent \textbf{Step 2:} Fix $ D, W, A $ and update $ P $:
				\begin{equation}\label{equation.8}
					P^{*} = {\operatorname*{argmin}_{P} \sum_{i=1}^{Q} \lambda_{1}\norm{P_{i}\overline{X_{i}}}_{F}^{2}+\lambda_{3}\norm{P_{i}X_{i}-A_{i}}_{F}^{2}}.
				\end{equation}
				We follow the same procedure as for $ A $ after solving the equation (\ref{equation.8}), $ P^{*} $ can be calculated with:
				\begin{equation}\label{equation.9}
					P^{*} = (\lambda_{3}X_{i}X_{i}^{T}+\lambda_{1}\overline{X_{i}} \: \overline{X_{i}}^{T}+\gamma I)^{-1}(\lambda_{3}A_{i}X_{i}^{T})
				\end{equation}
				where $ \gamma $ is a small number to prevent division by zero.
\noindent  \textbf{Step 3:} Fix $ P, D, A,$ and update $ W $
				with (\ref{equation.10}):
				\begin{equation}\label{equation.10}
					W_{i}^{*} = (A_{i}A_{i}^{T}+\gamma I)^{-1}(H_{i}A_{i}^{T}).
				\end{equation}

\noindent \textbf{Step 4:} Fix $ P, A, W $ and update $ D $:
				We obtain $D$ by using Alternating Direction Method of Multipliers (ADMM) algorithm \cite{goldstein2014fast} is as follows:
				
\begin{eqnarray}\label{equation.11}
			D^{(r+1)} &=& \operatorname*{min}_{D} \sum_{i=1}^{Q}\norm{X_{i}-D_{i}A_{i}}_{F}^{2}+\rho \norm{D_{i}-S_{i}^{(r)}+T_{i}^{(r)}}_{F}^{2}\nonumber \\	S^{(r+1)} &=& \operatorname*{min}_{S} \sum_{i=1}^{Q} \rho \norm{D_{i}^{(r+1)}-S_{i}^{(r)}+T_{i}^{(r)}}_{F}^{2} \nonumber\\ && st. ~ \norm{S_{i}} \le 1 \nonumber\\
		T^{(r+1)}&=&T^{(r)}+D_{i}^{(r+1)}-S_{i}^{(r+1)}.
	\end{eqnarray}
		
The pseudo-code of the proposed LpDPL approach is summarised in Algorithm \ref{Algorithm.1}. 

	\begin{algorithm}[t]
		\DontPrintSemicolon
		\SetAlgoLined
		\SetKwInOut{Input}{Input}\SetKwInOut{Output}{Output}
		\Input{Training samples for $Q$ classes $X=[X_{1},...,X_{2},..., X_{Q}]$, $m$, $\lambda_{1}$, $\lambda_{2}$, $\lambda_{3}$, $\gamma$ }
		Initialise $D_{0}$ and $P_{0}$ as random matrix and calculate $A_{0}$ in equation (\ref{equation.7}) and $W_{0}$ in equation (\ref{equation.10}), t = 0 \;
		
		\While{not converge}{
			t = t + 1\;
			\For{$i=1:k$}{
				Update $A^{(t)}_{k}\;$ by equation (\ref{equation.7})\;
				Update $P^{(t)}_{k}\;$ by equation (\ref{equation.9})\;
				Update $W^{(t)}_{k}\;$ by equation (\ref{equation.10})\;
				Update $D^{(t)}_{k}\;$ by equation (\ref{equation.11})\;
			}
		}
		\caption{Proposed LpDPL pseudo-code.}
		\Output{$P$, $D$, $W$}
		\label{Algorithm.1}
	\end{algorithm}
	
\subsection{Classification}
		Upon the completion of training with the labeled data in the proposed dictionary learning method, we obtain the learned synthesis and analysis dictionaries in addition to the transformation matrix associated to every class. Using $ P^{*} $, $ D^{*} $ and $ W^{*} $ from the training stage, a class label for testing a typical input image $x$ in vectorised form can obtained via:
		
		\begin{eqnarray}\label{equation.12}
			Class(x) &= &\operatorname*{argmin}_{i}~ \norm{x-D_{i}P_{i}}_{F}^{2}\nonumber\\&&+~\norm{H_{i}-W_{i}P_{i}x}_{F}^{2}.
		\end{eqnarray}			
\section{Experimental results}
\label{S:4}

We conducted extensive experiments to evaluate the effectiveness of the proposed method, and the corresponding results are presented in this section. The classification results of proposed method on two independent handwritten Chinese numbers databases are presented. We also evaluate the performance of the proposed method on an Arabic numbers database. Then, we compare the results of our proposed method with those obtained from deep learning architectures.

In all experiments, $ m, \lambda_{1} , \lambda_{2} $ and $\lambda_{3} $ were independently obtained using 10-fold cross-validation. We employed random initialisation for both $ D $ and $ P $ for each class. Then, these parameters are used to compute the initial $ A_{0} $ in equation (\ref{equation.7}) and consequently $ W_{0} $ in equation (\ref{equation.10}). 

\subsection{Case studies I \& II: Chinese number classification}

We computed the performance of the proposed method with using three different validation approaches: conventional, between-subjects and within-subjects. In conventional cross validation, we applied 10-fold cross validation for all images in the database. In the between-subject cross validation, we considered all data from one person as test set and data of all other persons were used as training set. We repeated this process for each participants handwriting. In the within-subject cross validation, we considered $n$-th sample from all the participants or people for testing set and the remaining samples were used as training set. This process was repeated 10 times.
Table \ref{table.1} shows the obtained results for the proposed method on Chinese handwritten numbers database. From this table, we can see our proposed method outperforms (3.8\%, in average) result reported in \cite{abolghasemi2018incoherent} where an incoherent DPL (InDPL) was used. Our proposed penalty terms, i.e. classification labels, in addition to new HOG features, have enhanced the performance of our dictionaries. To investigate the influence of using HOG features in the proposed method, we ran our method under the same experimental environment, i.e., parameters and data, however, without using the HOG features. The achieved classification accuracy reduced by $\sim11\%$. This experiment highlighted the significant impact of HOG features in this context.

\begin{table}[b]
	\caption{Classification accuracy of LpDPL and InDPL \cite{abolghasemi2018incoherent} for Chinese numbers classification. The results are presented for conventional, within-subject, and between-subject cross-validation.}
	\vspace{-.15cm}
\begin{center}
		\begin{tabular}{p{1.55cm}|c|c|c}
			Method & Conventional & Within-subject  & Between-subject \\
			\hline
			\hline
			LpDPL    &$ 98.53\% $&$98.56 \%$ & $98.07\% $ \\
			InDPL \cite{abolghasemi2018incoherent}  &$ 93.00\% $&$93.13 \% $ &$ 97.53\% $\\				
		\end{tabular}
	\end{center}
	\label{table.1}
\end{table}

\begin{figure*}[htp!]
	\begin{center}
		\includegraphics[scale=.37]{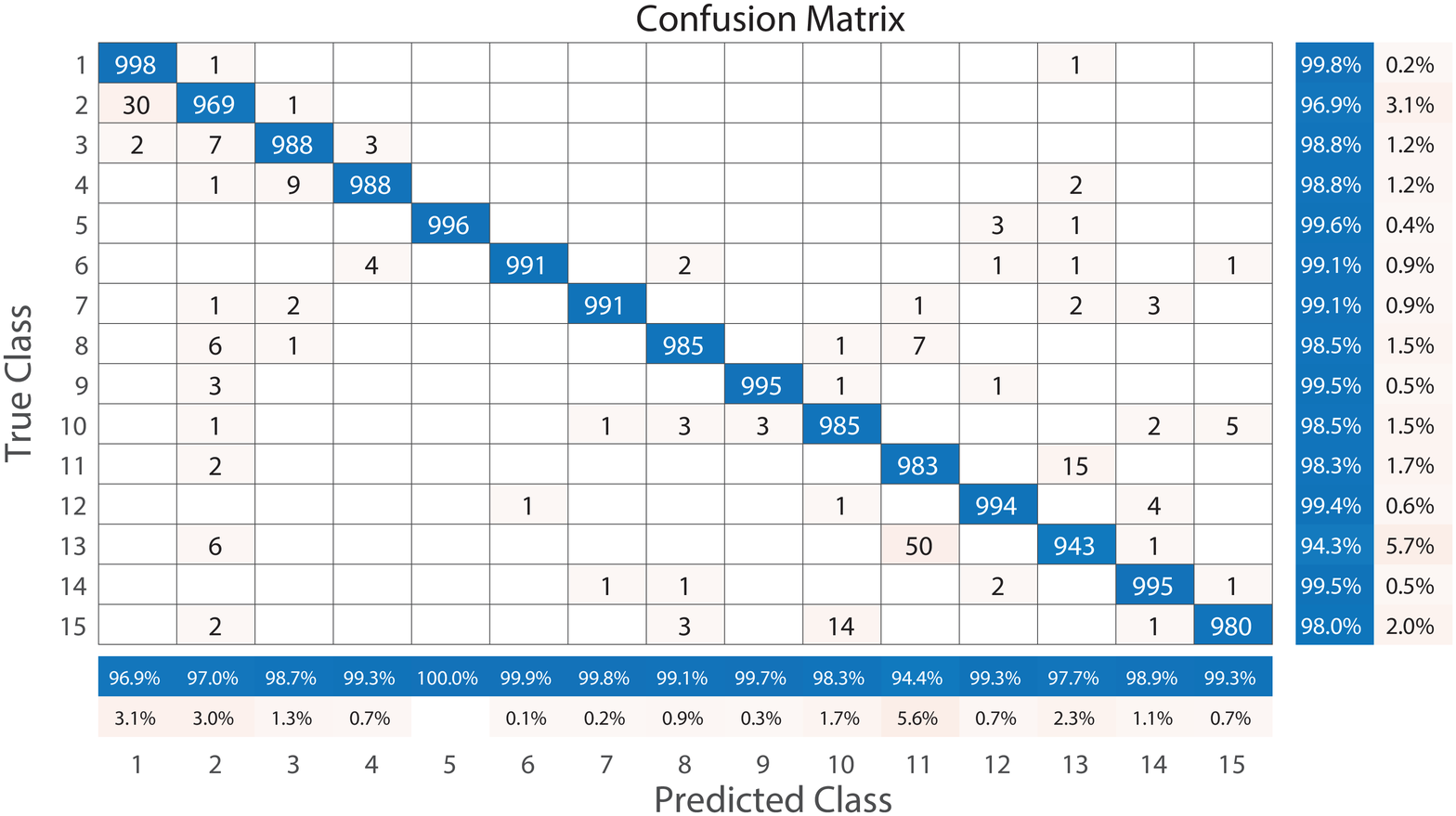}
		\caption{Confusion matrix representation as a result of performing conventional cross-validation experiment using LpDPL for handwritten Chinese database. Horizontal and vertical axis show the target (true) and output classes, respectively. The diagonal values show correct classification accuracy, and off-diagonals indicate misclassification associated to each target class.}\label{Figure.5}
	\end{center}
\end{figure*}

\begin{figure}[t]
	\centering \includegraphics[scale=.4]{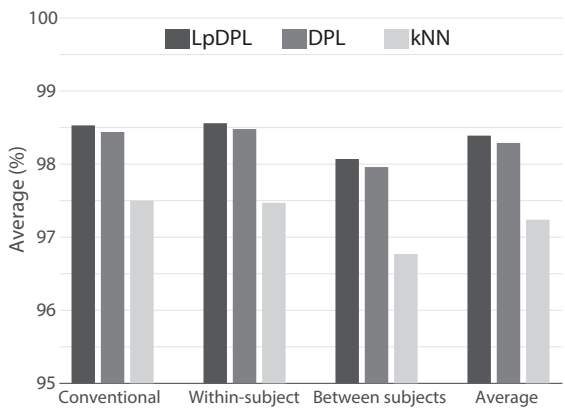}
			\caption{Comparison of classification accuracy for three methods, namely, LpDPL, DPL, and kNN, using different cross-validation settings.}\label{Fig_05}
\end{figure}

Figure \ref{Figure.5} shows the confusion matrix on Chinese handwritten numbers where it is notable that misclassification occurred between classes of ‘13’ and ‘11’ that refers to ‘1000’ and ‘10’ from Chinese numbers, respectively. This is due to the semantical similarities between these digits.
Next, we compare the proposed method with other classifiers such as k-nearest neighbor (kNN) and the original DPL~\cite{gu2014projective}. The comparison results are shown in Figure \ref{Fig_05}. The accuracy of classification using the proposed method here is higher compared to using same features.

We also compare the performance of the proposed method with existing dictionary leaning methods such as SRC \cite{wright2008robust}, DLSI \cite{ramirez2010classification}, LC-KSVD1 \cite{jiang2013label} and LC-KSVD2 \cite{jiang2013label} under conventional validation setting. In this analysis, we used two independent Chinese handwritten databases described in Section 2. As seen from the results in Table \ref{table.2}, the proposed method outperforms other well-established techniques. This reveals the effectiveness of the proposed hierarchy, i.e., the combination of the added penalty terms and HOG features.
\begin{table}
	\caption{Comparison of classification accuracy for popular dictionary learning methods on two Chinese handwritten databases; simplified and traditional.}
	\begin{center}
		\begin{tabular}{p{2.3cm}|p{2.5cm}|p{2.7cm}}
			Method & Simplified database & Traditional database \\
			\hline
			\hline			
			SRC	\cite{wright2008robust}			    	& $ 96.28 \%$ & $ 93.47 \%$ \\
			DLSI \cite{ramirez2010classification}		& $ 97.80 \%$ & $ 97.57 \%$ \\
			LC-KSVD1 \cite{jiang2013label}			& $ 95.23 \%$ & $ 90.65 \%$ \\
			LC-KSVD2 \cite{jiang2013label}			& $ 95.24 \%$ & $ 90.67 \%$
			\\
			LpDPL   								& $ 98.53 \%$ & $ 97.82 \%$ 
		\end{tabular}
	\end{center}
	\label{table.2}
\end{table}

\subsection{LpDPL versus deep learning} 
For completeness, we compared our method with well-framed deep learning models. In this experiment, we selected three well-established platforms, namely, GoogLeNet \cite{sandler2018mobilenetv2}, MobileNetV2 \cite{szegedy2015going}, and SqueezeNet \cite{iandola2016squeezenet}. To maximise the performance of these models, we used the fully-optimised version of the above models that were pre-trained on the very large ImageNet database \cite{deng2009imagenet}. The overal performance of LpDPL was $98.53\%$ which is comparable to GoogleNet ($ 99.83\%$), MobileNetV2 ($98.55\%$), and SqueezeNet ($98.53 \%$). The results also show that our method is more robust in recognising complex Chinese digits. e.g., number 9 and number 12; compared to CNN-based models. In the supplementary materials, we provide further details regarding class performances of each of the used deep learning methods.  

\subsection{Optimisation performance}
\begin{figure}[t]
	\centering \includegraphics[scale=0.5]{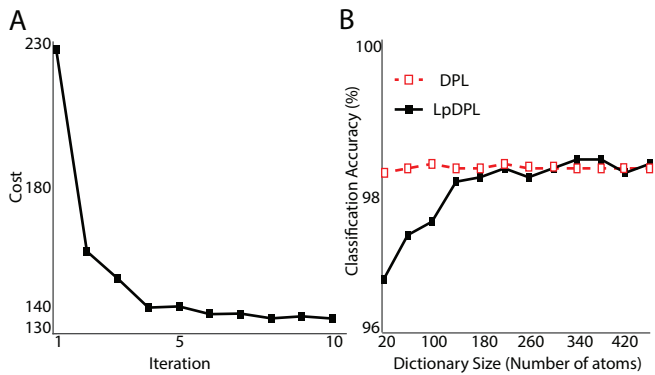}
	\caption{Optimisation performance; A) Cost function value (5) with respect to number of iterations; B) Classification accuracy against number of dictionary atoms.}\label{Fig_06}
\end{figure}

Figure \ref{Fig_06}A shows the optimisation process of objective function values for 10 iterations. The value of cost function in (5) against the number of iterations are represented in this graph. As expected, the objective function value decreases monotonically and quickly. 

To evaluate the effect of dictionary size on overall performance, we conducted an experiment with conventional 10-fold cross validation running at pre-defined dictionary sizes. Figure \ref{Fig_06}B shows the results of classification accuracy for proposed method and DPL, against different dictionary sizes (number of atoms $m$) for $D$. Based on the above analysis and due to the method consistent performance against various dictionary dimensions, we set $m=340$ for all of the experiments.

\subsection{Case study III: Arabic numbers classification}
As reviewed in Section I, HOG features have recently shown to be suitable descriptors for Arabic words too \cite{9092067}. Therefore, we expected the proposed method performs well with Arabic handwritten numbers as it relies on HOG features for dictionary learning and classification. 
In order to explore the performance of the proposed method, we used a publicly available Arabic number database (MADBase) \cite{AHD} as described in section II-A. We applied three dictionary learning methods, i.e. DPL, LpDPL, and SRC to this database and the results are given in Table \ref{table.4}. This table also reports the result of a previous work in \cite{ArabicCNN} where a CNN (LeNet-5) has applied to this data. Among all these methods, LpDPL achieved higher accuracy (Table \ref{table.4}). This experiment confirms the generalisation of the proposed method for handwritten databases in other languages. Another interesting finding that can be revealed by comparing results of Tables \ref{table.4} and \ref{table.1} is that HOG features has more tangible effects on Chinese numbers (than Arabic) which have complicated textures involving many line orientations. It is also noteworthy to mention that no significant changes in parameters were required for applying LpDPL to Arabic numbers.
\begin{table}[h]
	\caption{Classification results on Arabic handwritten numbers (MADBase).}
	\begin{center}
		\begin{tabular}{p{2cm}|c}
				Method & Accuracy (\%)\\
				\hline
				\hline			
				SRC	\cite{wright2008robust}			    	 & $ 97.13 $ \\
				LeNet-5	\cite{ArabicCNN}			    	 & $ 88.00 $ \\				
				DPL  \cite{gu2014projective}  				 & $ 98.25 $ \\
				LpDPL								 & $ 98.75 $ \\
			\end{tabular}
		\end{center}
		\label{table.4}
	\end{table}

\subsection{Case study IV: English numbers classification}
To further evaluate system performance, we applied the proposed method on an English handwritten numbers database. According to the obtained results with USPS database \cite{Hull1994}, represented in Table \ref{table.5}, the proposed method outperforms existing relevant techniques where highest average accuracy of 97.17 \% has achieved for LpDPL.
\begin{table}[h]
	\caption{Classification results on English handwritten numbers (USPS).}
	\begin{center}
		\begin{tabular}{p{2.5cm}|c}
			Method & Accuracy (\%)\\
			\hline
			\hline			
			SRC	\cite{wright2008robust}			    	& $ 81.81 $ \\
 			DLSI \cite{ramirez2010classification}		& $ 96.13 $ \\
 			LC-KSVD1 \cite{jiang2011learning}			& $ 91.25 $ \\
 			LC-KSVD2 \cite{jiang2011learning}			& $ 91.10 $ \\
 			DPL  \cite{gu2014projective}  				& $ 96.68 $ \\
 			LpDPL										& $ 97.17 $ \\
		\end{tabular}
	\end{center}
	\label{table.5}
\end{table}

\subsection{Parameter sensitivity study}
\begin{figure*}[t!]
	\centering
	\begin{subfigure}[b]{0.24\linewidth}
		\centering
		\includegraphics[width=\textwidth]{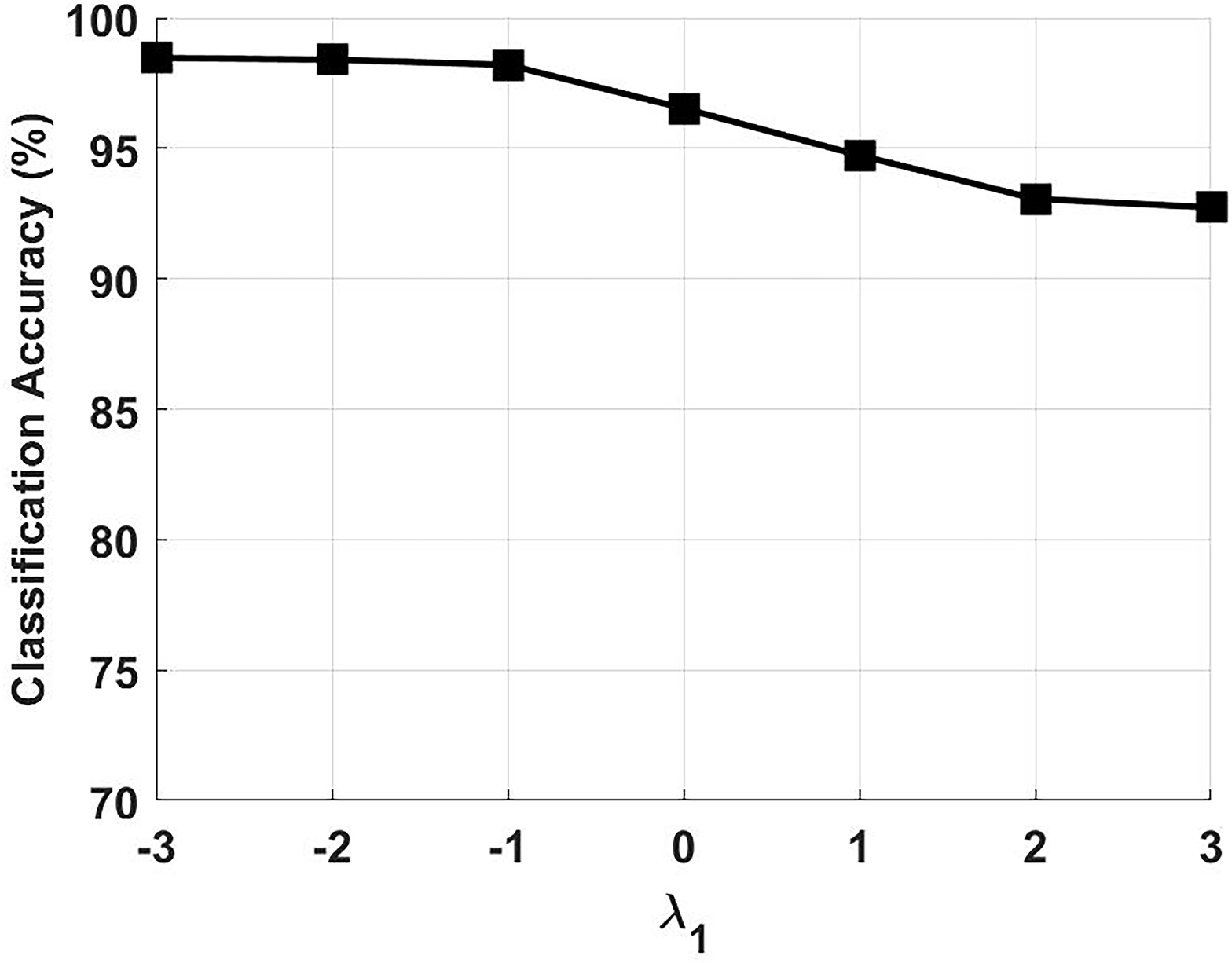}
		\caption{}
		\label{Fig_07A}	
	\end{subfigure}
	\hfill
	\begin{subfigure}[b]{0.24\linewidth}
		\centering
		\includegraphics[width=\textwidth]{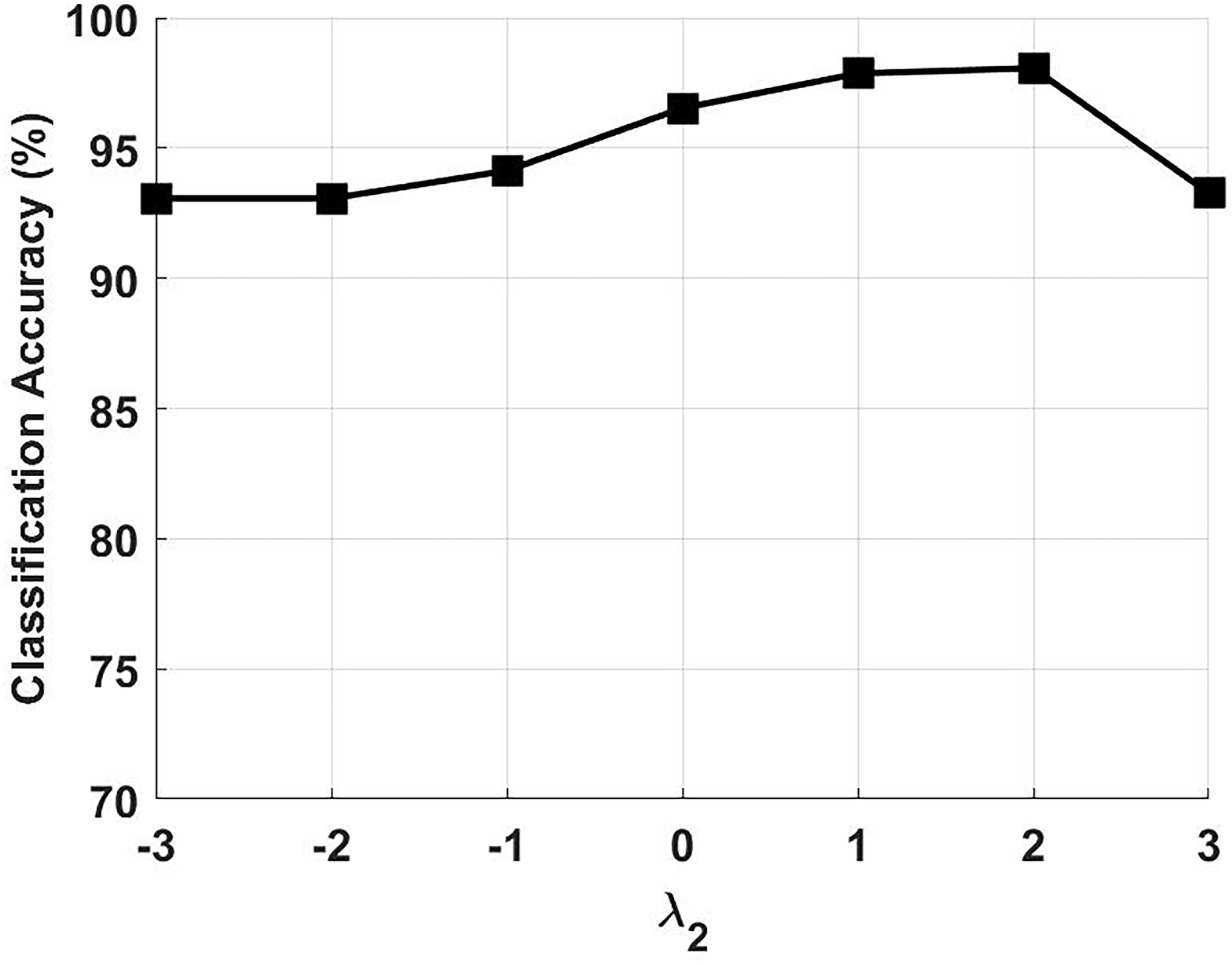}
		\caption{}
		\label{Fig_07B}	
	\end{subfigure}
	\hfill
	\begin{subfigure}[b]{0.24\linewidth}
		\centering
		\includegraphics[width=\textwidth]{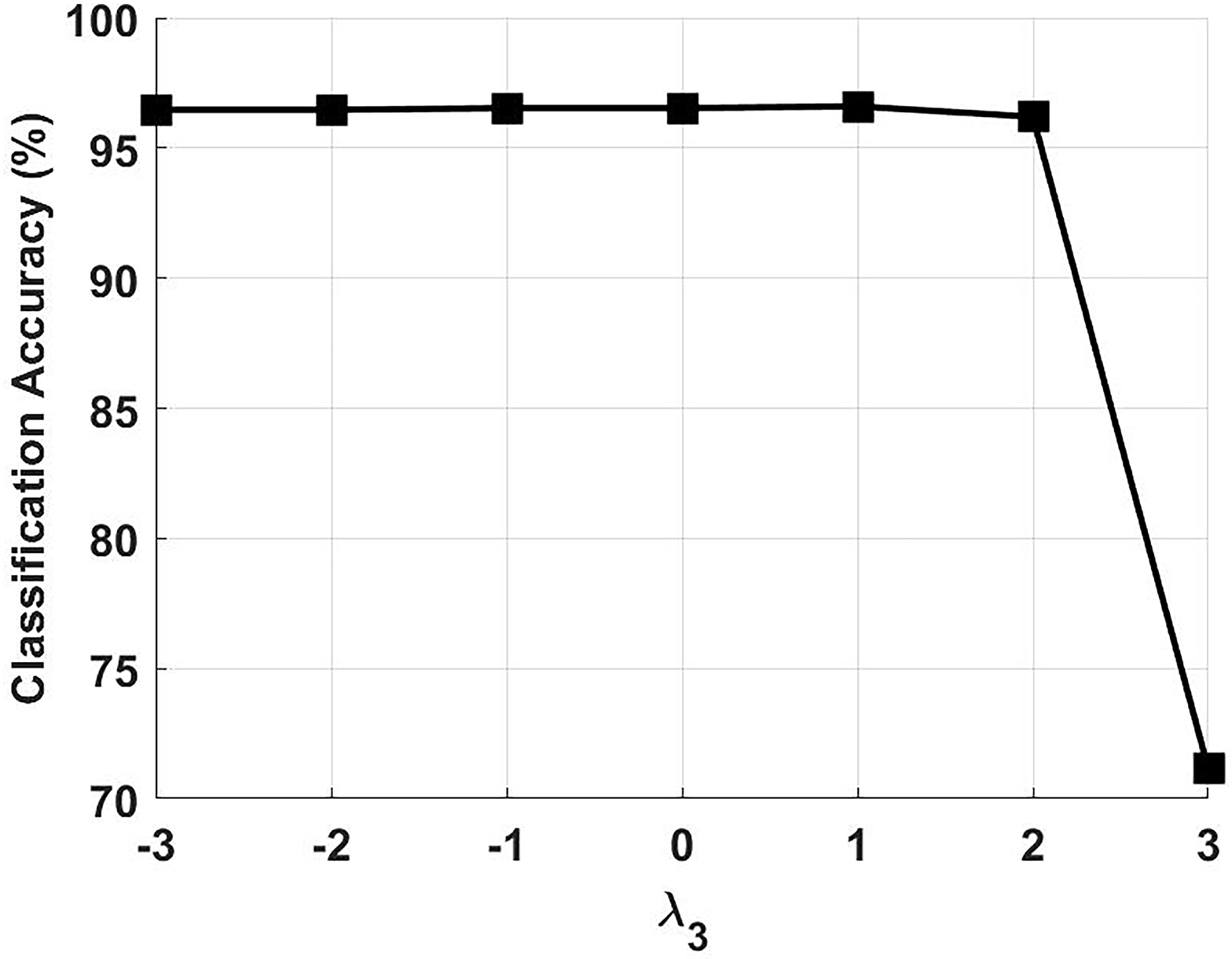}
		\caption{}
		\label{Fig_07C}	
	\end{subfigure}
	\hfill
	\begin{subfigure}[b]{0.24\linewidth}
		\centering
		\includegraphics[width=\textwidth]{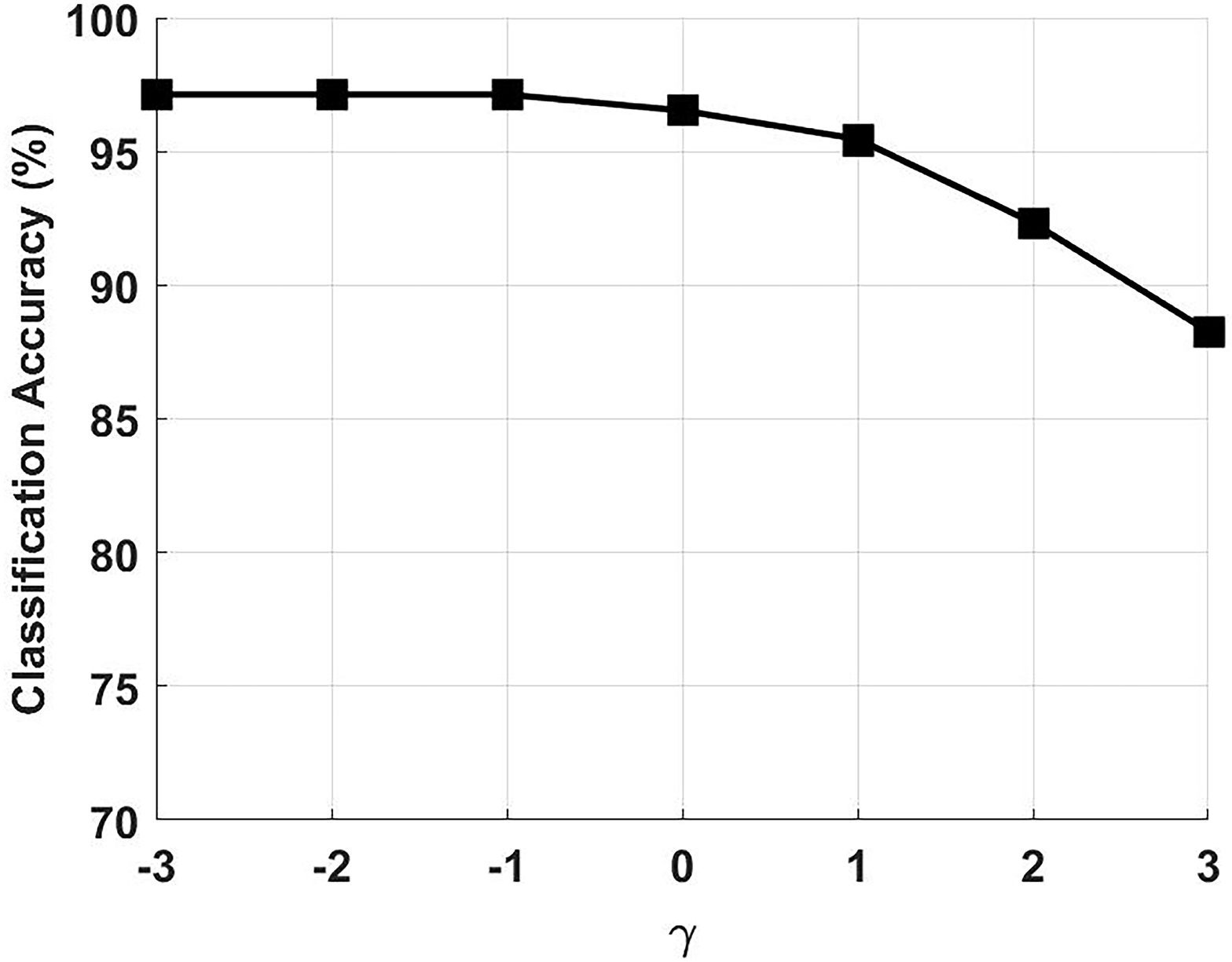}
		\caption{}
		\label{Fig_07D}	
	\end{subfigure}
\caption{Classification accuracy (\%) of LpDPL versus variations of the parameters (A) $\lambda_{1}$, (B) $\lambda_{2}$, (C) $\lambda_{3}$, (D) $\gamma$.}
\label{Fig_07}
\end{figure*}

\begin{figure*}[t!]
	\centering
	\begin{subfigure}[b]{0.3\linewidth}
		\centering
		\includegraphics[width=\textwidth]{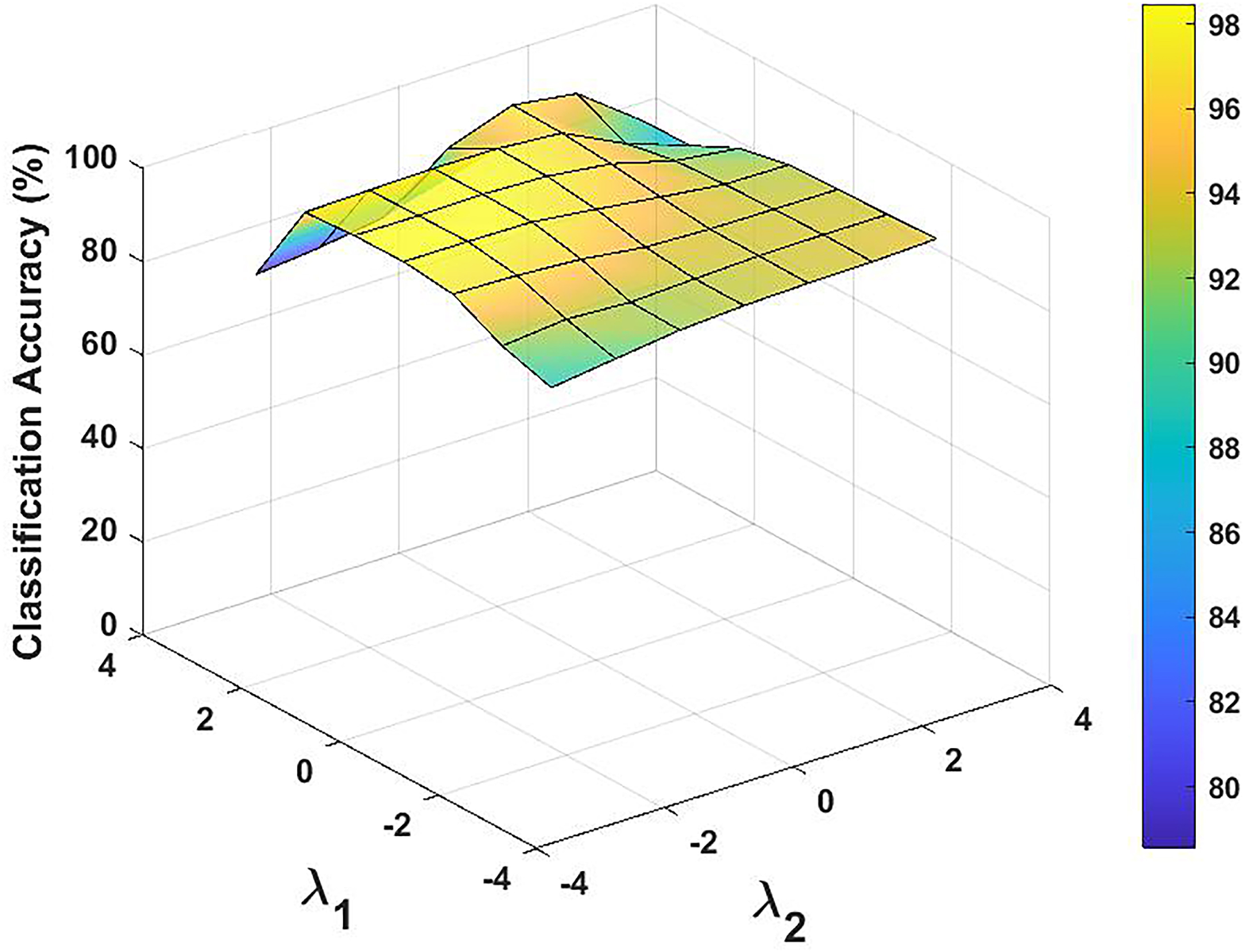}
		\caption{}
		\label{Fig_08A}	
	\end{subfigure}
	\hfill
	\begin{subfigure}[b]{0.3\linewidth}
		\centering
		\includegraphics[width=\textwidth]{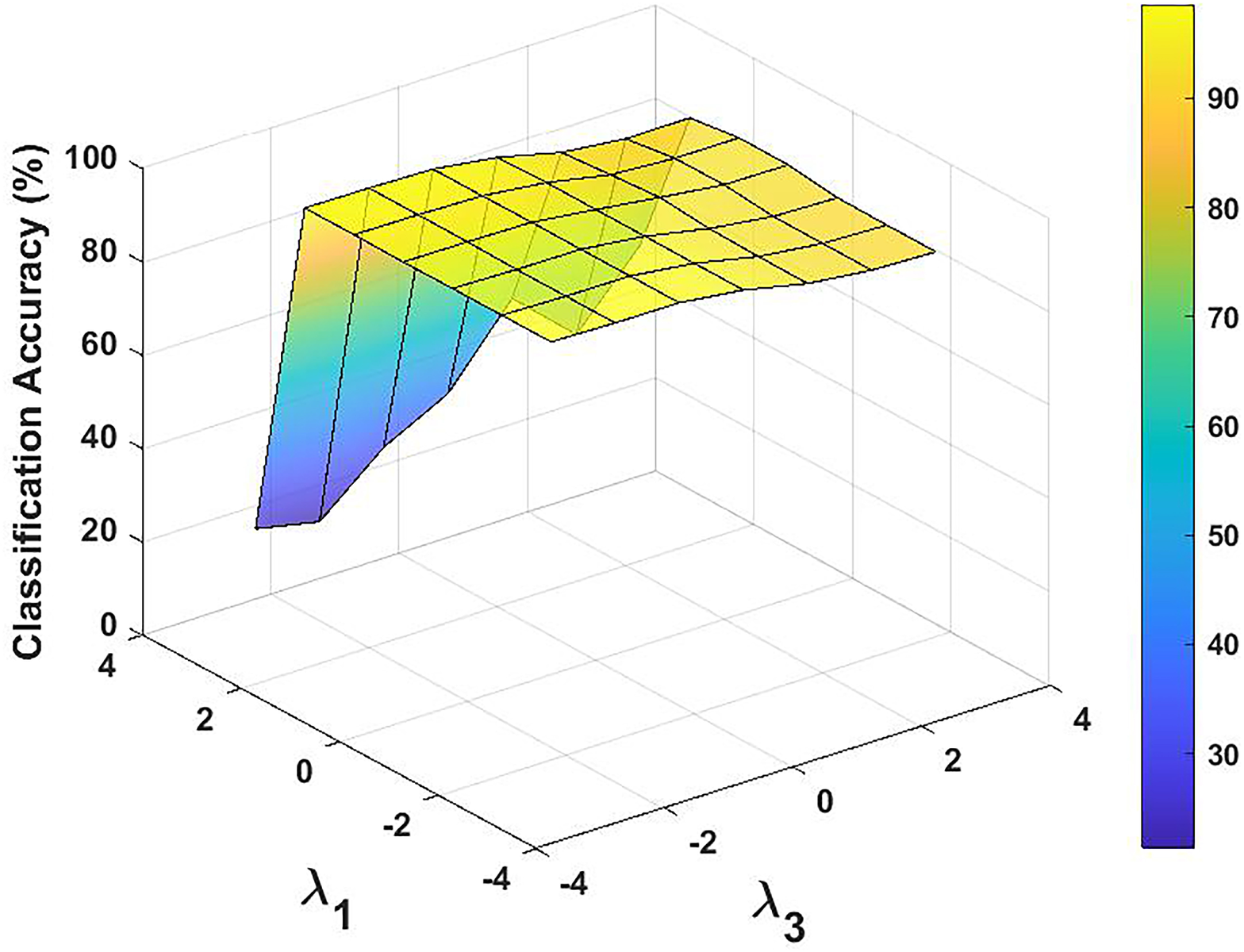}
		\caption{}
		\label{Fig_08B}	
	\end{subfigure}
	\hfill
	\begin{subfigure}[b]{0.3\linewidth}
		\centering
		\includegraphics[width=\textwidth]{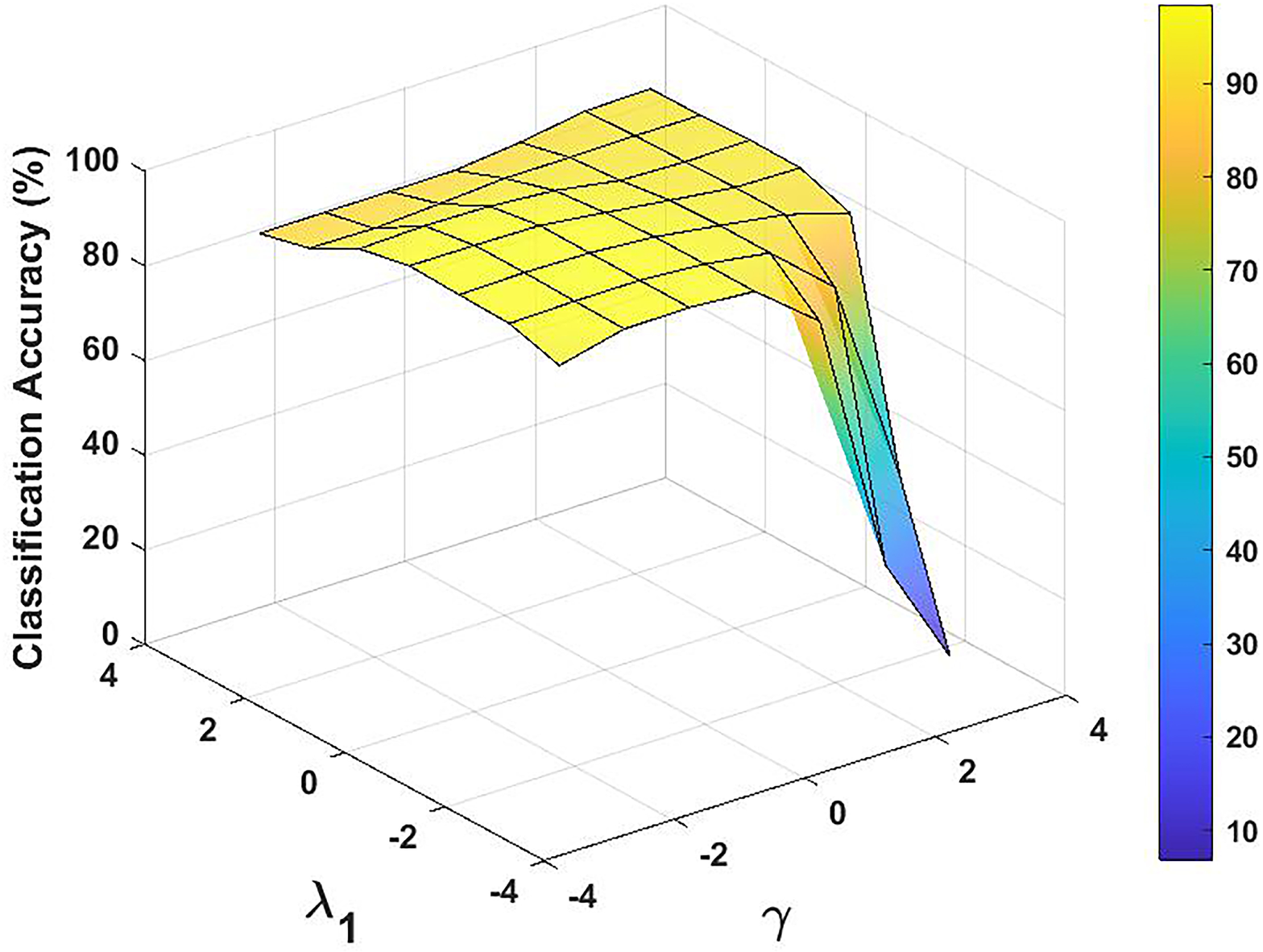}
		\caption{}
		\label{Fig_08C}	
	\end{subfigure}
	\newline
		\begin{subfigure}[b]{0.3\linewidth}
		\centering
		\includegraphics[width=\textwidth]{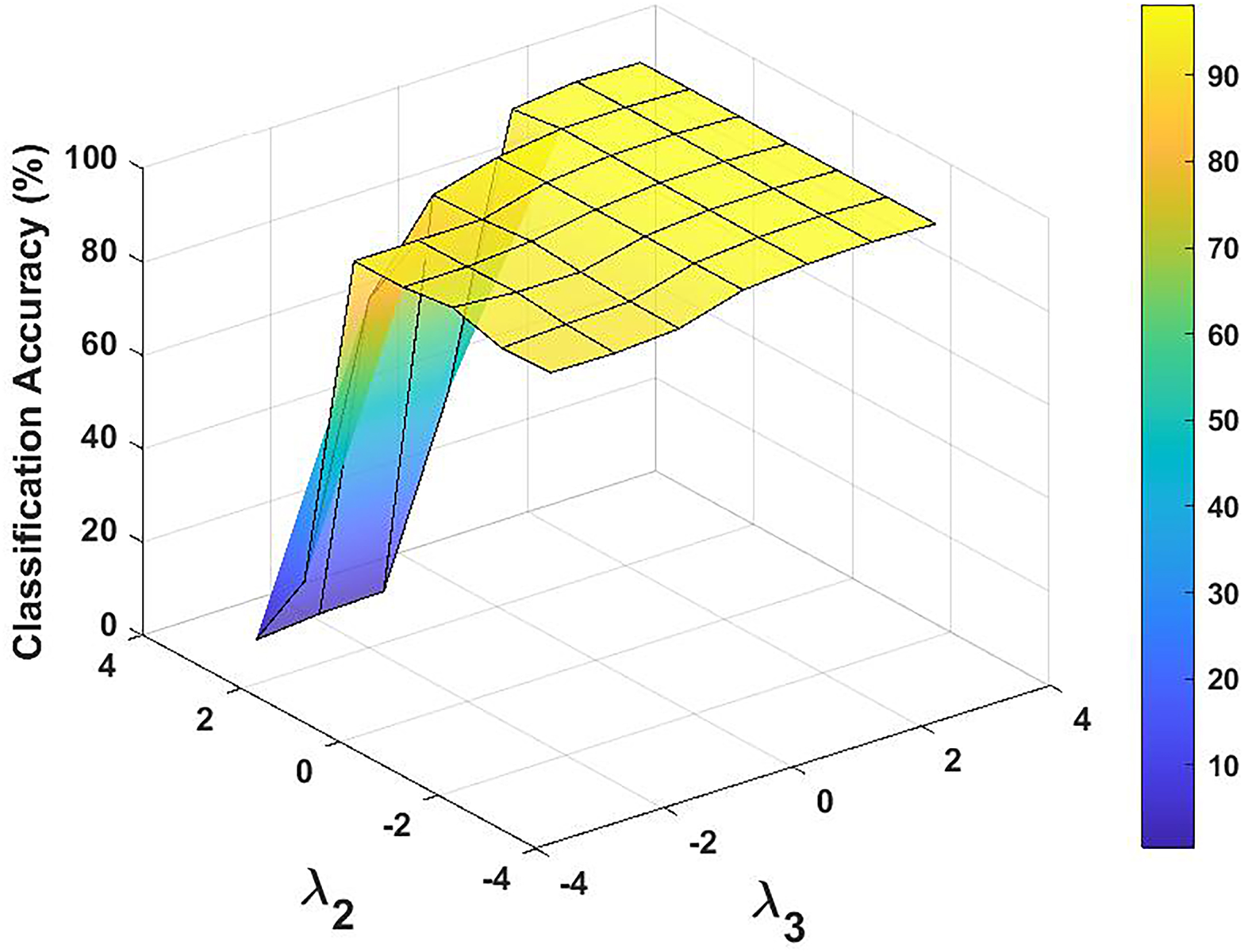}
		\caption{}
		\label{Fig_08D}	
	\end{subfigure}
	\hfill
	\begin{subfigure}[b]{0.3\linewidth}
		\centering
		\includegraphics[width=\textwidth]{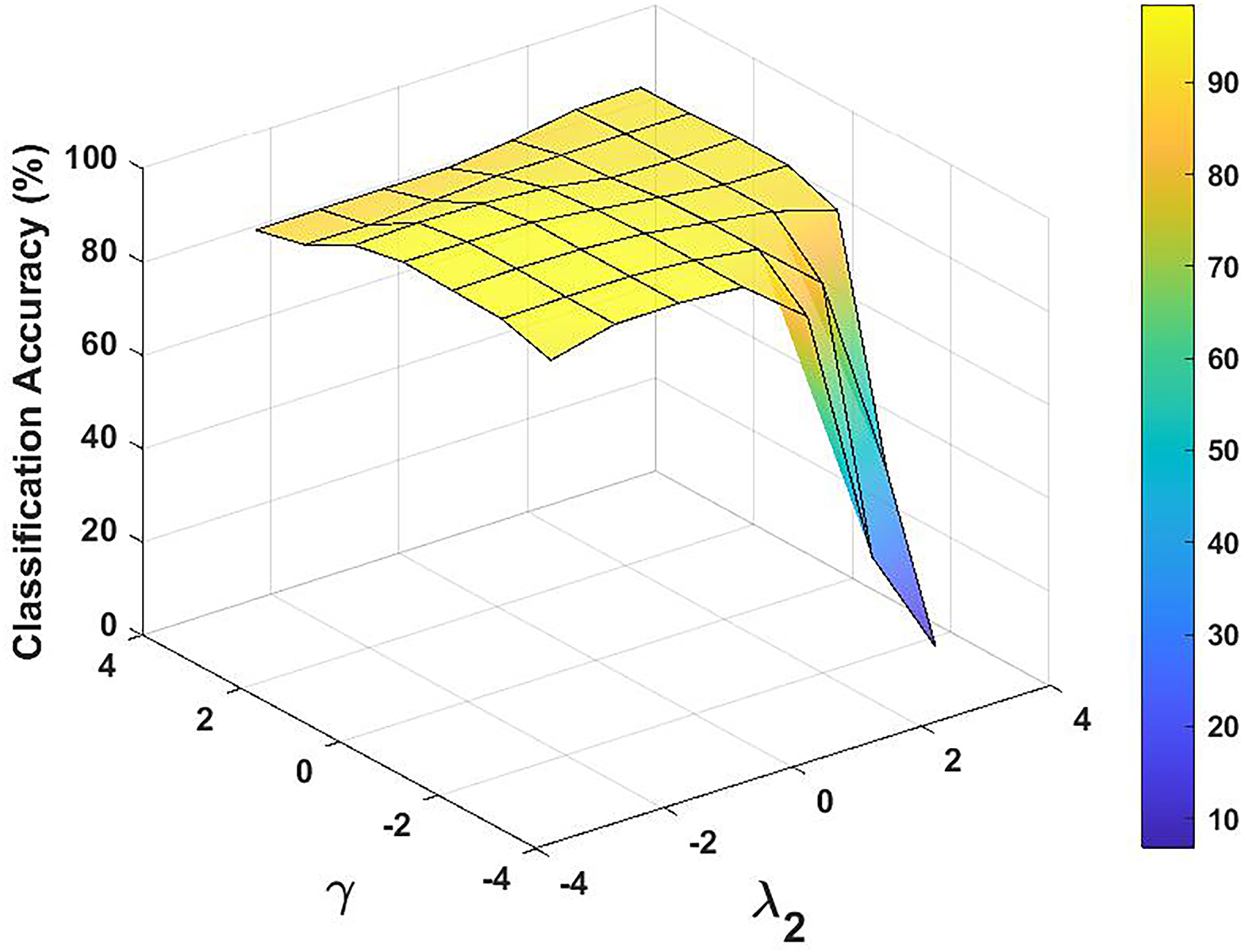}
		\caption{}
		\label{Fig_08E}	
	\end{subfigure}
	\hfill
	\begin{subfigure}[b]{0.3\linewidth}
		\centering
		\includegraphics[width=\textwidth]{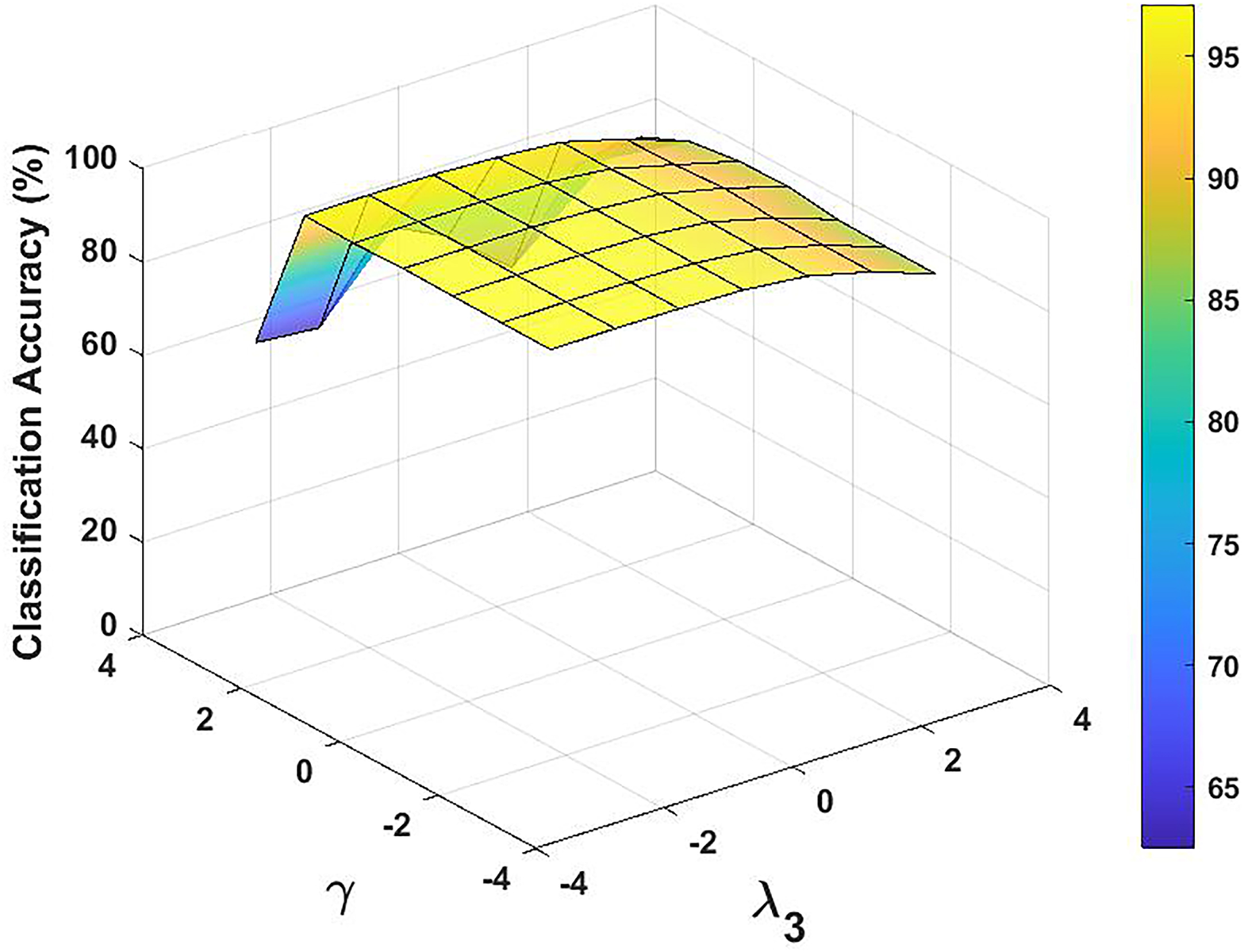}
		\caption{}
		\label{Fig_08F}	
	\end{subfigure}
	\caption{Classification accuracy (\%) of LpDPL versus variations of the parameters (A) $\lambda_{1}$, $\lambda_{2}$, (B) $\lambda_{1}$, $\lambda_{3}$, (C) $\lambda_{1}$, $\gamma$, (D) $\lambda_{2}$, $\lambda_{3}$, (E) $\gamma$, $\lambda_{2}$, and (F) $\gamma$, $\lambda_{3}$.}
	\label{Fig_08}
\end{figure*}

In order to assess the robustness of the proposed method, we record the recognition performance of LpDPL over the variations of key parameters in Algorithm 1, i.e., $\lambda_{1}$, $\lambda_{2}$, $\lambda_{3}$ and $\gamma$. For this purpose, at each experiment, we fine-tune the value of one parameter in the range $[10^{-3},10^3]$, while keeping other parameters fixed. Figure \ref{Fig_07} and Figure \ref{Fig_08} demonstrate the recognition accuracy (\%) of LpDPL versus variations of these parameters on simplified Chinese handwritten numbers. In particular, we have the following observations from these figures: When $\lambda_1>10^{-1}$, LpDPL slightly suffers a performance drop due to overweighting the discrimination factor. Interestingly, increasing contribution of class labels information (i.e. increasing $\lambda_2$) improves the performance. However, the performance drops for very large values, i.e., $\lambda_2>10^2$. LpDPL experiences a significant performance degradation when $\lambda_3>10^{2}$. We believe this is due to significant reduction of the contributions of discrimination power and class-label information when such a large $\lambda_2$ is selected. We also observe that when $\gamma>10^0$, LpDPL's performance starts to drop (Figure \ref{Fig_08D}). This is because $\gamma$ is purposed to avoid zero division in (\ref{equation.9}). Therefore, a large $\gamma$ leads to an inaccurate dictionary $P$. Overall, we observed that LpDPL is not sensitive to the parameters' variations within a broad range.

\subsection{On Real-Time Implementation}
The proposed methods were developed and implemented using MATLAB R2018a with Intel core i7 with 2.20 GHz processor and 8 GB of memory. The deep learning experiments were conducted on Ubuntu 18.04 with Matlab 2019b environment using NVIDIA GeForce RTX 2080 Ti. 
In a real-time setting, dictionaries may be trained and updated offline. With our method, the feature extraction and classification times are $0.64$ms and $0.24$ms per image, respectively, without relying on expensive GPUs. This allows our model to run using embedded low-computational computers, for example on a Raspberry PI.

\section{Conclusions}
We proposed a new labeled projective dictionary pair learning approach. Unlike most existing dictionary learning methods which use $\ell_0$-norm and $\ell_1$-norm to calculate sparse code, our approach is able to calculate sparse code by linear projection. More importantly, we utilised HOG features into the dictionary learning hierarchy and added available class labels as penalty term into the cost function. We tested it with classifying two Chinese handwritten numbers databases in addition to an Arabic handwritten numbers database. The experimental results show that our approach yields excellent classification performance that were higher than that with conventional methods. Unlike deep learning methods, our method runs on computers with modest specifications; it runs all the data locally and it does not require GPU devices or cloud processing; two standard mechanisms for running deep learning models. Finally, GoogLeNet, MobileNetV2, and SqueezeNet, require 7, 3.5 and 1.24 million parameters respectively; the proposed model requires the fine-tunning of only eight parameters.

As the future work, we are going to explore the possibility of combining deep learning and dictionary learning (particularly DPL due to using a pair of synthesis-analysis dictionaries) for the purpose of Chinese handwritten numbers recognition.

\balance

\bibliographystyle{IEEEtran}
\bibliography{ref}

\end{document}